\pdfoutput=1
\documentclass{article}

    \PassOptionsToPackage{numbers, compress}{natbib}

\usepackage[preprint]{tsrml_2022}

\usepackage[utf8]{inputenc} 
\usepackage{url}            
\usepackage{booktabs}       
\usepackage{amsfonts}       
\usepackage{nicefrac}       
\usepackage{microtype}      
\usepackage{xcolor}         
\usepackage{amsmath}
\usepackage{verbatim}
\usepackage{booktabs}
\usepackage{multirow}
\usepackage{siunitx}

\usepackage{amsmath}
\usepackage{amssymb}       
\usepackage{amsthm} 
\usepackage{wrapfig}

\usepackage{graphicx}
\usepackage{float}
\usepackage{epstopdf}
\usepackage{esvect}
\usepackage{caption, subcaption}
\usepackage{geometry}
\usepackage{xcolor}
\usepackage{wrapfig}
\newcounter{todocounter}

\usepackage{empheq} 
\usepackage{xcolor}
\definecolor{lightgreen}{HTML}{90EE90}

\usepackage{breqn}

\title{Scoring Black-Box Models for Adversarial Robustness}

\author{%
  Jian Vora, Pranay Reddy Samala\\
  Department of Computer Science, Stanford University\\
  \texttt{\{jianv, pranayr\}@cs.stanford.edu} \\
}

\begin{document}

\maketitle

\begin{abstract}

Deep neural networks are susceptible to adversarial inputs and various methods have been proposed to defend these models against adversarial attacks under different perturbation models. The robustness of models to adversarial attacks has been analyzed by first constructing adversarial inputs for the model, and then testing the model performance on the constructed adversarial inputs. Most of these attacks require the model to be white-box, need access to data labels and finding adversarial inputs can be computationally expensive. We propose a simple scoring method for black-box models which indicates their robustness to adversarial input. We show that adversarially more robust models have a smaller $l_1$-norm of \textsc{Lime} weights and sharper explanations. 
\end{abstract}

\section{Introduction}
Deep neural networks have shown impressive performance on a variety of tasks in domains such as vision \cite{dosovitskiy2020image, pmlr-v139-ramesh21a}, natural language \cite{NEURIPS2020_1457c0d6}, speech \cite{Hsu2021HuBERTSS, baevski2020wav2vec}. As and when we start deploying these models in the real world, it becomes important that these models are truly robust and reliable. One way is to have \textit{model cards} \footnote{\texttt{https://modelcards.withgoogle.com/}}\cite{mitchell2019model} associated with models which not only include their downstream performance metric but also important aspects such as robustness \cite{goodfellow_harnessing}, fairness \cite{fairness}, training datasets, etc. Such model cards would provide practitioners a much better insight into model selection for a particular application leading to the safer deployment of these models in the real world. Among other issues with current deep learning models, one crucial concern while deploying these models is their brittleness to specially designed adversarial inputs that fool them. In this work, we aim at scoring black-box models based on their robustness to adversarial samples. We score models based on the explanations they generate which leads to an interesting connection between robustness and explainability of models. In the upcoming subsections, we provide a brief overview of adversarial robustness and input attribution-based explanation methods for machine learning models. 

\subsection{Adversarial Robustness}
\label{adv_robust}
Current deep learning models have been shown to be vulnerable to human-imperceptible adversarial perturbations which significantly degrade the model's performance. This is not desirable, especially in safety-critical applications such as autonomous driving \cite{xu2022safebench} and medical applications \cite{finlayson2018adversarial}. While the theory of adversarial robustness is quite general, we restrict the discussion here to image classifier models. Most classes of adversarial attacks try to add small perturbations to the input so as to move the input sample across a decision boundary, thus changing the model's prediction. Various types of attacks have been tried to target classifiers followed by various defense mechanisms as well. A common theme for image-classifier based attack methods is to find a perturbation $\boldsymbol{\delta}$ to be added to an input $\mathbf{x} \in \mathbb{R}^D$ such that $\tilde{\mathbf{x}} = \mathbf{x} + \boldsymbol{\delta}$ fools the classifier. Different constraints of the perturbation $\boldsymbol{\delta}$ lead to different types of attacks, the most common of which is trying to bound some $l_p$ norm of $\boldsymbol{\delta}$. Concretely, if $L$ is the loss and $\mathcal{M}$ is the the neural network, all these attacks try to find the perturbation $\boldsymbol{\delta}$ by solving the following optimization problem,
\begin{equation}
\label{eqn: adv_equation}
    \max_{\boldsymbol{\delta}} L(\mathcal{M}(\mathbf{x} + \boldsymbol{\delta})) \text{   s.t.   } \|\boldsymbol{\delta}\|_p \leq \epsilon
\end{equation}

White-box attacks are the most common ones in which the adversary has complete access to the model parameters and details of any defense mechanism. Some common white-box attacks include Fast Gradient Sign Method (FGSM) \cite{goodfellow_harnessing}, Carlini \& Wagner \cite{carlini2017towards}, DeepFool \cite{moosavi2016deepfool}, L-BFGS \cite{szegedy2013intriguing}, JSMA \cite{jsma}. The mainstream approach to defend against these attacks is by training on adversarial examples \cite{goodfellow_harnessing, kurakin2017adversarial, madry2018towards, Shaham2015, na2018cascade, tramer2018ensemble}. Other methods include defense distillation \cite{defensive_distillation}, and input transformations to make inputs closer to the training distribution (e.g. via randomization or generative modelling) \cite{pmlr-v89-zhang19b, samangouei2018defensegan, Yoon2021AdversarialPW, nie2022diffusion, song2018pixeldefend}. 

\subsection{Explainability}
Many methods have been proposed that try to provide post-hoc explanations of predictions from a black-box machine learning model \cite{ribeiro2016should, shap}. For a given input sample, all these methods try to attribute parts of the input that led to a particular prediction from the model. In this work, we utilize \textsc{Lime} \cite{ribeiro2016should}, a popular method for explanations. Linear functions are inherently explainable with the coefficients giving the relative importance of a feature attribute in predicting the outcome. \textsc{Lime} uses this fact to approximate a model as a linear function of the input around the query data points.

Concretely, given a black-box model $\mathcal{M}$ trained on input samples $\{\mathbf{x}_i\}_{i=1}^{N}$ with $\mathbf{x}_i \in \mathcal{X} \subseteq \mathbb{R}^d$ generating predictions $\hat{y}_i = \mathcal{M}(\mathbf{x}_i)$, \textsc{Lime} does the following to generate explanations for query sample $\mathbf{x_i}$:
\begin{enumerate}
    \item Sample $k$ points near $\mathbf{x}_i$ and let them be called $\{\mathbf{z}_{i, j}\}_{j=1}^{k}$
    \item Fit a linear model $\mathbf{w}_i^T\mathbf{z}_i (\mathbf{w}_i \in \mathbb{R}^d)$ on a new dataset $\{\mathbf{z}_{i, j}, \mathcal{M}(\mathbf{z}_{i, j}) \}_{j=1}^k$
\end{enumerate}
We refer the reader to \cite{ribeiro2016should} for different methods of sampling near $\mathbf{x}_i$ for Step $1$. These methods mainly depend on the domain where $\mathbf{x}_i$ belongs (for example images, text, etc.). Step 2 is simply performing linear regression on the sampled points around $\mathbf{x}_i$ and predictions from the model $\mathcal{M}$ on those points. The learnt weights $\mathbf{w}_i \in \mathbb{R}^d$ can be used to explain the input sample $\mathbf{x}_i$. These weights can be thought of as approximating the model $M$ with a linear model around $\mathbf{x}_i$ and indicate how the model predictions will change given small changes in $\mathbf{x}_i$. In the subsequent sections, we shall denote these linear weights $\mathbf{w}_i$ as $\textsc{Lime}(\mathcal{M}(\mathbf{x}_i))$. An important point to note here is that computing $\textsc{Lime}(\mathcal{M}(\mathbf{x}_i))$ does not require access to the model weights and can be done for any black-box model $\mathcal{M}$.
\section{Method}

In this section, we shall define our scoring method for any black-box model based on the adversarial robustness of the model.
Typically, robustness is measured by first generating adversarial inputs to the model and then measuring the performance of the models on the generated adversarial inputs. For a model $\mathcal{M}$, we shall denote \texttt{robust-acc}$(\mathcal{M})$ as the accuracy of model $\mathcal{M}$ on adversarial inputs generated by one of the perturbation models described in Section ~\ref{adv_robust}. Most common adversarial attacks are white-box, i.e. they need the exact model weights to generated adversarial inputs. Given a perturbation model and two models $\mathcal{M}_1$ and $\mathcal{M}_2$, we call $\mathcal{M}_1$ to be more robust then $\mathcal{M}_2$ \textit{iff} \texttt{robust-acc}($\mathcal{M}_1$) $>$ \texttt{robust-acc}($\mathcal{M}_2$).

We note that the above way of comparing methods for robustness has certain issues:
\begin{enumerate}
    \item Most methods which generate adversarial inputs require model weights. However, we note that many current models are available as usable APIs which makes them black-box. It is hard to find to adversarial inputs for black-box models which makes it hard to compute \texttt{robust-acc} and subsequently compare models for robustness.
    \item To compute \texttt{robust-acc}, we need access to labels of the input samples to compare with model predictions which might not always be available.
    \item Even for white-box models, computing \texttt{robust-acc} is computationally expensive. Finding a single adversarial sample requires performing several steps of gradient ascent on the input space to solve Eq. \ref{eqn: adv_equation}.
\end{enumerate}
We propose a scoring method which does not have the above limitations. The method works for any black-box model off-the-shelf, does not require finding adversarial samples and works for any unlabeled dataset. Under these constraints, we validate that our scoring method correlates well with \texttt{robust-acc}. More formally, the only information we have access to is an unlabelled dataset $\mathcal{X}$, a trained black-box model $\mathcal{M}$.

Let $\mathcal{X}$ be the space of inputs and $\mathcal{Y}$ be the output space. $\mathcal{M}_1$ and $\mathcal{M}_2$ are two black-box models mapping $\mathcal{X} \to \mathcal{Y}$. Let ${\bf x}_1, {\bf x}_2, ... {\bf x}_n$ denote $n$ sample points drawn from $\mathcal{X}$ and we do not have associated labels $y$ for any of them. Denoting lime weights for model input ${\bf x}$ by $\textsc{Lime}(\mathcal{M}({\bf x}))$, input dataset size by $n$, and data dimensionality $d$, we propose,
\begin{dmath}[frame]
\label{brittle-score}
\texttt{brittle-score}_{\mathcal{X}}(\mathcal{M}) = \frac{1}{nd}\sum_{{\bf x}_i \in \mathcal{X}} \|\textsc{Lime}(\mathcal{M}({\bf x}_i))\|_1
\end{dmath}

where $\|.\|_1$ denotes the vector $l_1$ norm. \texttt{brittle-score} indicates the model's \textit{`brittleness'} to adversarial inputs and hence is inversely related to \texttt{robust-acc}. Note that the above score only depends on the black-box model $\mathcal{M}$ and input dataset $\mathcal{X}$. Our \textit{hypothesis} is that,
\begin{center}
 $\mathcal{M}_1$ is more robust than $\mathcal{M}_2$ $\Leftrightarrow $ \texttt{robust-acc}($\mathcal{M}_1$) $>$ \texttt{robust-acc}($\mathcal{M}_2$)
 \\ \hspace{5.15cm}  $\Leftrightarrow $ \texttt{brittle-score}($\mathcal{M}_1$) $<$ \texttt{brittle-score}($\mathcal{M}_2$)  
\end{center}
The intuition is simple -- more robust models should have the property that small changes in input do not affect the output a lot. Let  $\nabla_{\bf x}\mathcal{M}(\bf x)$ denote the gradient of the model output with respect to the input $\mathbf{x}$. We expect that the adversarially robust models will have a lower norm of $\nabla_{\bf x}\mathcal{M}(\bf x)$. As the models are black-box, we cannot compute gradients with respect to the input and hence use $\textsc{Lime}(\mathcal{M}({\bf x})) \approx \nabla_{\bf x}M(\bf x)$ as a proxy. The intuition behind this is similar to prior work that explains adversarial examples for linear models by Goodfellow et al. \cite{goodfellow_harnessing}. 

Note that the value of \texttt{brittle-score} as defined in Eq. \ref{brittle-score} is simply the magnitude of the lime weights averaged over $n$ points, hence this value is not directly interpretable (unlike \texttt{robust-acc}). However, the real purpose of this score is to \textit{compare} the adversarial robustness across two black-box models. Also note that the above scoring method is specific to a dataset $\mathcal{X}$ and should not be compared across models trained on different datasets. In the subsequent section, we present experimental results which show that \texttt{brittle-score} is a very good indicator of the robustness of a model and correlates inversely with \texttt{robust-acc}.

\section{Experiments}
In this section, we demonstrate how adversarial training affects both \texttt{robust-acc} and \texttt{brittle-score}. The main takeaway from the results presented in this Section is that \texttt{brittle-score} is inversely correlated to \texttt{robust-acc} for a variety of model architectures and attacks and hence makes it an attractive scoring method for black-box models.

\begin{figure}
    \centering
    \includegraphics[scale=0.45]{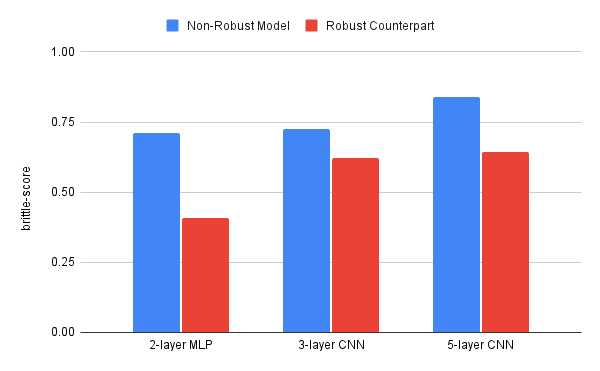}
    \caption{Comparison of \texttt{brittle-score} of various types of models trained normally or in an adversarial manner. Red bar being consistently lower than blue validates our hypothesis.}
    \label{fig:placeholder}
\end{figure}

We first define some controlled experiments on MNIST \cite{lecunmnist} dataset. Our first set of experiments involves adversarial training over different model architectures. We experiment with three different model architectures -- $2$-layer MLP, a $3$-layer CNN and a $5$-layer CNN. For each of these architectures, we train two models:
\begin{enumerate}
    \item One without adversarial training trained on samples from the data distribution using a cross-entropy loss with learning rate $10^{-3}$ and a batch size of $32$.
    \item One with adversarial training where the samples were generated by attacking each model with PGD \cite{madry2018towards} attack under $l_\infty$ constraint ($p = \infty$ in Eq. \ref{eqn: adv_equation}) of $\epsilon = {8}/{255}$ for $\boldsymbol{\delta}$.
\end{enumerate}
In Figure \ref{fig:placeholder}, we plot the \texttt{brittle-score} (Eq. \ref{brittle-score}) computed on $n=1000$ samples for the each of the models. These samples were drawn from the data distribution (validation split in this case) and were not \textit{attack samples}. We observe a significant difference in \texttt{brittle-score} between robust and non-robust models with the robust models having a lower value of \texttt{brittle-score}. 

\begin{figure*}[!h]
\centering
\begin{subfigure}{.48\textwidth}
  \centering
  \includegraphics[width=.99\linewidth]{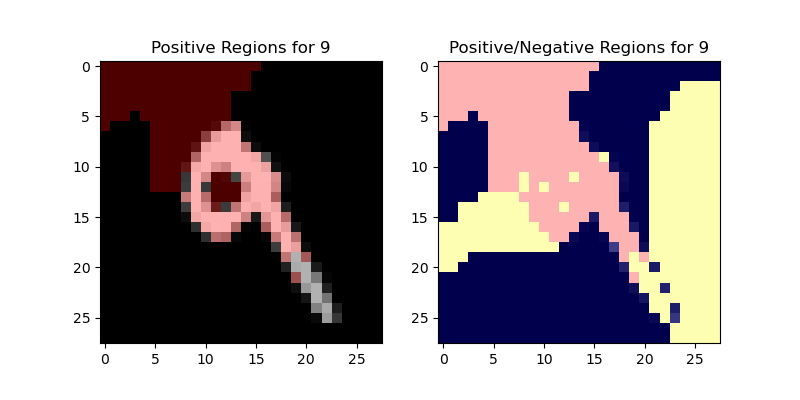}
\end{subfigure}
\begin{subfigure}{.48\textwidth}
  \centering
  \includegraphics[width=0.99\linewidth]{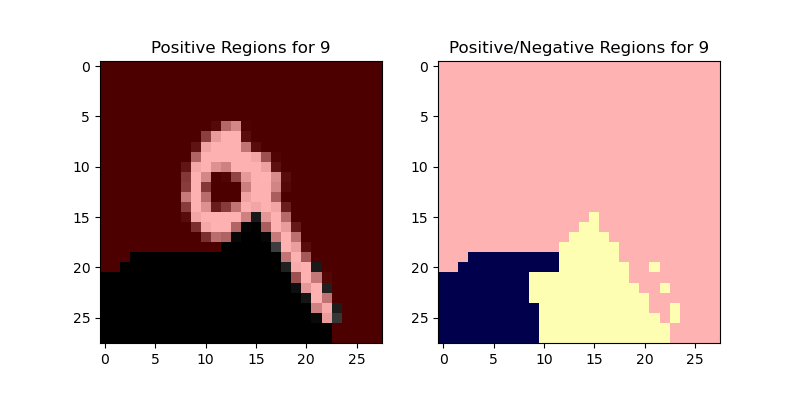}
\end{subfigure}%
\caption{(Left) Explanations of robust 3-layer CNN and (Right) Explanations of non-robust 3-layer CNN. The robust model has better lime weights (sparser positive region) which is seen in larger yellow colored regions in the mask (blue=negative, pink=positive, yellow=zero) }
\label{fig:adv9}
\end{figure*}

We also look at the explanations generated by these models qualitatively. We notice that explanations generated by the more robust model are typically sharper and focus on the most relevant parts of the input images (Fig. \ref{fig:adv9}).
For non-robust models, the \textsc{Lime} weights are not as sparse. This in-turn motivates the definition of \texttt{brittle-score} which depends on the hypothesis that \textsc{Lime} weights of a robust model has a lower norm than the non-robust counterpart.

\begin{figure*}[!h]
\centering
\begin{subfigure}{.48\textwidth}
  \centering
  \includegraphics[width=.99\linewidth]{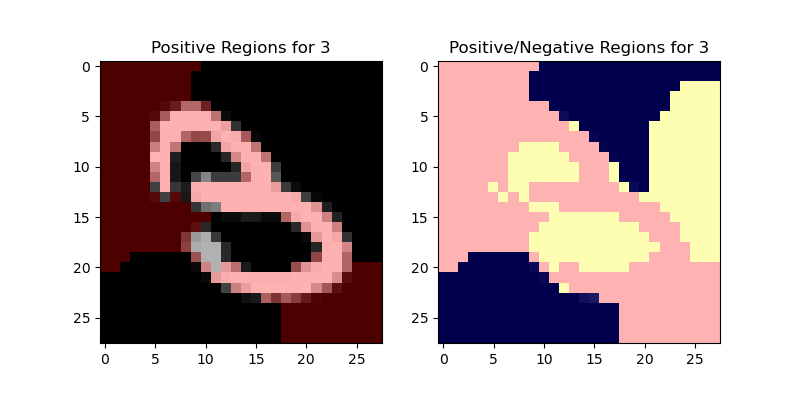}
\end{subfigure}
\begin{subfigure}{.48\textwidth}
  \centering
  \includegraphics[width=0.99\linewidth]{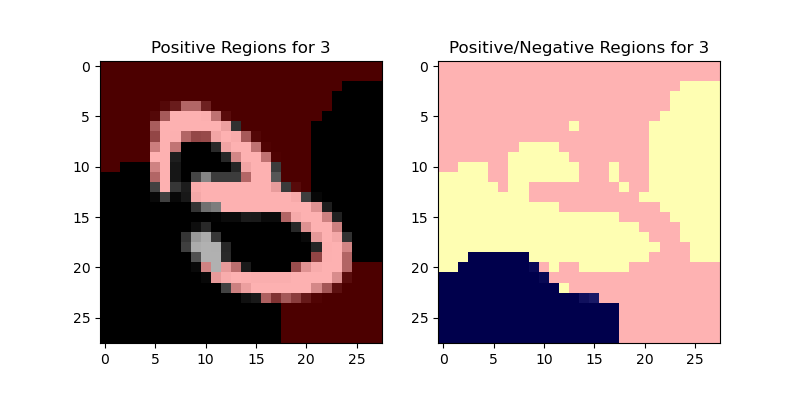}
\end{subfigure}%
\label{fig:adv3}
\end{figure*}

In addition to our controlled studies on MNIST, we also perform experiments on off-the-shelf open-source adversarially trained models available as a part of \texttt{robustbench}\footnote{\texttt{https://github.com/RobustBench/robustbench}}\cite{croce2021robustbench}. We test with three other datasets -- CIFAR10, CIFAR100 and ImageNet. \texttt{robustbench} provides a bunch of pre-trained checkpoints on these datasets which were defended using different methods and hence they also have different robust accuracies. This gives us a good way to compare \texttt{robust-acc} and \texttt{brittle-score} in order to validate our hypothesis. We use $n=3000$ in Eq. \ref{brittle-score} for CIFAR10 and CIFAR100 and $n=100$ for ImageNet with $k=1000$ samples for fitting the \textsc{Lime} function around a datapoint. 

\begin{figure}[!h]
\centering
\begin{subfigure}{.49\textwidth}
  \centering
  \includegraphics[width=.99\linewidth]{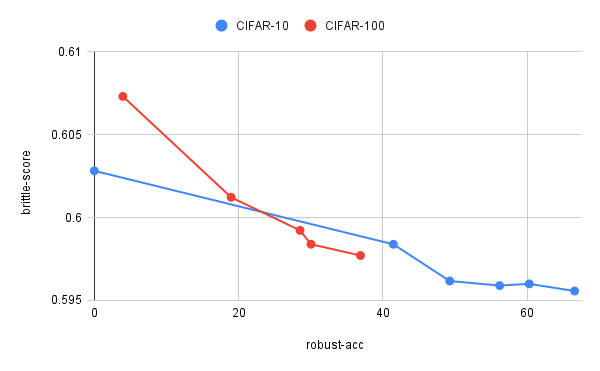}
\end{subfigure}
\begin{subfigure}{.49\textwidth}
  \centering
  \includegraphics[width=0.99\linewidth]{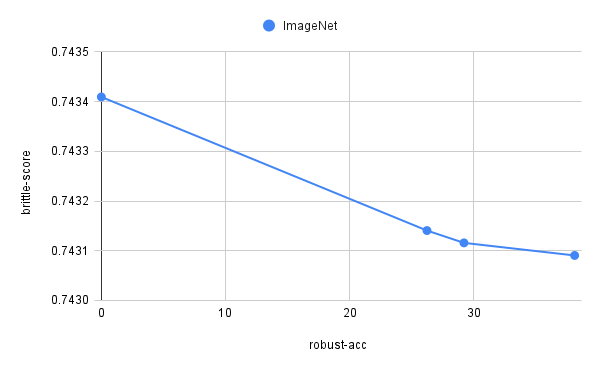}
  
\end{subfigure}%
\caption{We plot \texttt{brittle-score} versus \texttt{robust-acc} for various datasets based on Table \ref{tab:modelvbrittle} showing that both of them are negatively correlated to one another}
\label{fig:my_label}
\end{figure}

In Table \ref{tab:modelvbrittle}, we show that with increasing \texttt{robust-acc}, the \texttt{brittle-score} decreases. The correlation between these two metrics is particularly intriguing since these two quantities prima facie appear to be completely unrelated. \texttt{robust-acc} is directly reported from \texttt{robustbench}\cite{croce2021robustbench}. We also report the relative improvement (R.I.) for a model $\mathcal{M}$ over the standard model denoted as $\mathcal{S.M}$ for each dataset which is defined as:
\begin{equation}
\label{eq: ri}
        \text{R.I.} = \frac{\texttt{brittle-score(}\mathcal{S.M}) - \texttt{brittle-score(}\mathcal{M})}{\texttt{brittle-score(}\mathcal{S.M})} \times 100
\end{equation}
\begin{table}[H]
    \centering
    \begin{tabular}{c c c c c} \toprule
        Dataset & Model Name & \texttt{robust-acc} & \texttt{brittle-score} & R.I. 
        \\\toprule 
         \multirow{6}{4em}{CIFAR-10} & Standard (Non-Robust) \cite{He2016DeepRL} & $0.0$ & $0.6028$ & $0$\\ 
         & Ding et al. \cite{Ding2020MMA} & $41.44$ & $0.5984$ & $0.74$\\ 
         & Engstrom et al. \cite{engstrom} & $49.25$ & $0.5962$ & $1.11$\\ 
         & Wu et al. \cite{wu2020adversarial} & $56.17$ & $0.5960$ & $1.14$\\ 
         & Sehwag et al. \cite{sehwag2022robust} & $60.27$ & $0.5960$ & $1.14$\\ 
         & Rebuffi et al. \cite{rebuffi2021data} & $\mathbf{66.56}$ &  $\mathbf{0.5956}$ & $\mathbf{1.22}$ \\ \midrule
         \multirow{6}{4em}{CIFAR-100} & Standard (Non-Robust) \cite{He2016DeepRL} & $3.95$ & $0.6073$ & $0$\\ 
         & Rice et al. \cite{Rice2020OverfittingIA} & $18.95$ & $0.6012$ & $1.00$\\ 
         & Rebuffi et al. \cite{rebuffi2021data} & $28.5$ & $0.5992$ & $1.34$\\ 
         & Gowal et al. \cite{gowaluncovering} & $30.03$ & $0.5984$ & $1.49$\\ 
         & Gowal et al. (Extra Model) \cite{gowaluncovering} & $\mathbf{36.88}$ & $\mathbf{0.5977}$ & $\mathbf{1.60}$\\ \midrule
         \multirow{6}{4em}{ImageNet} & Standard (Non-Robust) \cite{He2016DeepRL} & $0.0$ & $0.74341$ & $0$\\ 
         & Wong et al.\cite{Wong2020Fast} & $26.24$ & $0.74314$ & $0.036$\\ 
         & Engstrom et al. \cite{engstrom} & $29.22$ & $0.74312$ & $0.039$\\ 
         & Salman et al. \cite{salman} & $\mathbf{38.14}$ & $\mathbf{0.74309}$ & $\mathbf{0.043}$\\\midrule 
    \end{tabular}
    
    \caption{A comparison of the trend between robust accuracy and brittle score across model architectures, adversarial defenses and datasets. R.I. standards for relative improvement as defined in Eq. \ref{eq: ri}.}
    \label{tab:modelvbrittle}
\end{table}
As mentioned earlier, \texttt{brittle-score} should only be compared within datasets. The absolute value of \texttt{brittle-score} is not interpretable as it is but it serves as a good metric to compare models. If we try to compare two different models which were trained and tested on different datasets, the absolute numbers and differences do not have a mathematical meaning. Moreover, it is sometimes hard to interpret the scale of \texttt{brittle-score} by itself. Unlike \texttt{robust-acc}, the scale of \texttt{brittle-score} is unbounded, and could be any number $> 0$, although we empirically observe that the number is $< 1$. For robust models, the difference in \texttt{brittle-score} itself is constantly shrinking, indicating a sub-linear relationship between \texttt{robust-acc} and \texttt{brittle-score}. We plot \texttt{robust-acc} versus \texttt{brittle-score} in Fig. \ref{fig:my_label} as both of them should be negatively correlated.







\section{Conclusion \& Future Directions}

We illustrate that \texttt{brittle-score} is a great indicator of the robustness of any black-box model. We believe this is quite encouraging from not only a practical standpoint, but from a theoretical standpoint as well in understanding adversarial robustness and adversarial training. Moreover, this method makes evaluating and comparing black box models (for instance models provided via APIs), feasible and easy. With the increasing proliferation of deep learning methods, methods that can preserve intellectual property rights while indicating model safety are paramount for safety-centric deep learning. Some potential extensions of our work include: (a) a theoretical analysis to understand the relation between adversarial robustness and \textsc{Lime} weights, (b) developing black-box scoring methods that capture other safety related metrics.
\bibliographystyle{plain}
\bibliography{ref}

\begin{thebibliography}{10}

\bibitem{baevski2020wav2vec}
Alexei Baevski, Yuhao Zhou, Abdelrahman Mohamed, and Michael Auli.
\newblock wav2vec 2.0: A framework for self-supervised learning of speech
  representations.
\newblock {\em Advances in Neural Information Processing Systems},
  33:12449--12460, 2020.

\bibitem{NEURIPS2020_1457c0d6}
Tom Brown, Benjamin Mann, Nick Ryder, Melanie Subbiah, Jared~D Kaplan, Prafulla
  Dhariwal, Arvind Neelakantan, Pranav Shyam, Girish Sastry, Amanda Askell,
  Sandhini Agarwal, Ariel Herbert-Voss, Gretchen Krueger, Tom Henighan, Rewon
  Child, Aditya Ramesh, Daniel Ziegler, Jeffrey Wu, Clemens Winter, Chris
  Hesse, Mark Chen, Eric Sigler, Mateusz Litwin, Scott Gray, Benjamin Chess,
  Jack Clark, Christopher Berner, Sam McCandlish, Alec Radford, Ilya Sutskever,
  and Dario Amodei.
\newblock Language models are few-shot learners.
\newblock In H.~Larochelle, M.~Ranzato, R.~Hadsell, M.F. Balcan, and H.~Lin,
  editors, {\em Advances in Neural Information Processing Systems}, volume~33,
  pages 1877--1901. Curran Associates, Inc., 2020.

\bibitem{carlini2017towards}
Nicholas Carlini and David Wagner.
\newblock Towards evaluating the robustness of neural networks.
\newblock In {\em 2017 ieee symposium on security and privacy (sp)}, pages
  39--57. Ieee, 2017.

\bibitem{croce2021robustbench}
Francesco Croce, Maksym Andriushchenko, Vikash Sehwag, Edoardo Debenedetti,
  Nicolas Flammarion, Mung Chiang, Prateek Mittal, and Matthias Hein.
\newblock Robustbench: a standardized adversarial robustness benchmark.
\newblock In {\em Thirty-fifth Conference on Neural Information Processing
  Systems Datasets and Benchmarks Track}, 2021.

\bibitem{Ding2020MMA}
Gavin~Weiguang Ding, Yash Sharma, Kry Yik~Chau Lui, and Ruitong Huang.
\newblock Mma training: Direct input space margin maximization through
  adversarial training.
\newblock In {\em International Conference on Learning Representations}, 2020.

\bibitem{dosovitskiy2020image}
Alexey Dosovitskiy, Lucas Beyer, Alexander Kolesnikov, Dirk Weissenborn,
  Xiaohua Zhai, Thomas Unterthiner, Mostafa Dehghani, Matthias Minderer, Georg
  Heigold, Sylvain Gelly, et~al.
\newblock An image is worth 16x16 words: Transformers for image recognition at
  scale.
\newblock {\em arXiv preprint arXiv:2010.11929}, 2020.

\bibitem{engstrom}
Logan Engstrom, Andrew Ilyas, Hadi Salman, Shibani Santurkar, and Dimitris
  Tsipras.
\newblock Robustness (python library), 2019.

\bibitem{finlayson2018adversarial}
Samuel~G Finlayson, Hyung~Won Chung, Isaac~S Kohane, and Andrew~L Beam.
\newblock Adversarial attacks against medical deep learning systems.
\newblock {\em arXiv preprint arXiv:1804.05296}, 2018.

\bibitem{goodfellow_harnessing}
Ian~J. Goodfellow, Jonathon Shlens, and Christian Szegedy.
\newblock Explaining and harnessing adversarial examples.
\newblock In {\em 3rd International Conference on Learning Representations,
  {ICLR} 2015, San Diego, CA, USA, May 7-9, 2015, Conference Track
  Proceedings}, 2015.

\bibitem{gowaluncovering}
Sven Gowal, Chongli Qin, Jonathan Uesato, Timothy Mann, and Pushmeet Kohli.
\newblock Uncovering the limits of adversarial training against norm-bounded
  adversarial examples, 2020.

\bibitem{He2016DeepRL}
Kaiming He, X.~Zhang, Shaoqing Ren, and Jian Sun.
\newblock Deep residual learning for image recognition.
\newblock {\em 2016 IEEE Conference on Computer Vision and Pattern Recognition
  (CVPR)}, pages 770--778, 2016.

\bibitem{Hsu2021HuBERTSS}
Wei-Ning Hsu, Benjamin Bolte, Yao-Hung~Hubert Tsai, Kushal Lakhotia, Ruslan
  Salakhutdinov, and Abdelrahman Mohamed.
\newblock Hubert: Self-supervised speech representation learning by masked
  prediction of hidden units.
\newblock {\em IEEE/ACM Transactions on Audio, Speech, and Language
  Processing}, 29:3451--3460, 2021.

\bibitem{kurakin2017adversarial}
Alexey Kurakin, Ian~J. Goodfellow, and Samy Bengio.
\newblock Adversarial machine learning at scale.
\newblock In {\em International Conference on Learning Representations}, 2017.

\bibitem{lecunmnist}
Y~LECUN.
\newblock The mnist database of handwritten digits.
\newblock {\em http://yann. lecun. com/exdb/mnist/}, 1999.

\bibitem{shap}
Scott~M Lundberg and Su-In Lee.
\newblock A unified approach to interpreting model predictions.
\newblock In I.~Guyon, U.~V. Luxburg, S.~Bengio, H.~Wallach, R.~Fergus,
  S.~Vishwanathan, and R.~Garnett, editors, {\em Advances in Neural Information
  Processing Systems 30}, pages 4765--4774. Curran Associates, Inc., 2017.

\bibitem{madry2018towards}
Aleksander Madry, Aleksandar Makelov, Ludwig Schmidt, Dimitris Tsipras, and
  Adrian Vladu.
\newblock Towards deep learning models resistant to adversarial attacks.
\newblock In {\em International Conference on Learning Representations}, 2018.

\bibitem{fairness}
Ninareh Mehrabi, Fred Morstatter, Nripsuta Saxena, Kristina Lerman, and Aram
  Galstyan.
\newblock A survey on bias and fairness in machine learning.
\newblock {\em ACM Comput. Surv.}, 54(6), jul 2021.

\bibitem{mitchell2019model}
Margaret Mitchell, Simone Wu, Andrew Zaldivar, Parker Barnes, Lucy Vasserman,
  Ben Hutchinson, Elena Spitzer, Inioluwa~Deborah Raji, and Timnit Gebru.
\newblock Model cards for model reporting.
\newblock In {\em Proceedings of the conference on fairness, accountability,
  and transparency}, pages 220--229, 2019.

\bibitem{moosavi2016deepfool}
Seyed-Mohsen Moosavi-Dezfooli, Alhussein Fawzi, and Pascal Frossard.
\newblock Deepfool: a simple and accurate method to fool deep neural networks.
\newblock In {\em Proceedings of the IEEE conference on computer vision and
  pattern recognition}, pages 2574--2582, 2016.

\bibitem{na2018cascade}
Taesik Na, Jong~Hwan Ko, and Saibal Mukhopadhyay.
\newblock Cascade adversarial machine learning regularized with a unified
  embedding.
\newblock In {\em International Conference on Learning Representations}, 2018.

\bibitem{nie2022diffusion}
Weili Nie, Brandon Guo, Yujia Huang, Chaowei Xiao, Arash Vahdat, and Anima
  Anandkumar.
\newblock Diffusion models for adversarial purification.
\newblock In {\em ICML}, 2022.

\bibitem{jsma}
Nicolas Papernot, Patrick McDaniel, Somesh Jha, Matt Fredrikson, Z.~Berkay
  Celik, and Ananthram Swami.
\newblock The limitations of deep learning in adversarial settings.
\newblock In {\em 2016 IEEE European Symposium on Security and Privacy
  (EuroS\&P)}, pages 372--387, 2016.

\bibitem{defensive_distillation}
Nicolas Papernot, Patrick McDaniel, Xi~Wu, Somesh Jha, and Ananthram Swami.
\newblock Distillation as a defense to adversarial perturbations against deep
  neural networks.
\newblock In {\em 2016 IEEE Symposium on Security and Privacy (SP)}, pages
  582--597, 2016.

\bibitem{pmlr-v139-ramesh21a}
Aditya Ramesh, Mikhail Pavlov, Gabriel Goh, Scott Gray, Chelsea Voss, Alec
  Radford, Mark Chen, and Ilya Sutskever.
\newblock Zero-shot text-to-image generation.
\newblock In Marina Meila and Tong Zhang, editors, {\em Proceedings of the 38th
  International Conference on Machine Learning}, volume 139 of {\em Proceedings
  of Machine Learning Research}, pages 8821--8831. PMLR, 18--24 Jul 2021.

\bibitem{rebuffi2021data}
Sylvestre-Alvise Rebuffi, Sven Gowal, Dan~Andrei Calian, Florian Stimberg,
  Olivia Wiles, and Timothy Mann.
\newblock Data augmentation can improve robustness.
\newblock In A.~Beygelzimer, Y.~Dauphin, P.~Liang, and J.~Wortman Vaughan,
  editors, {\em Advances in Neural Information Processing Systems}, 2021.

\bibitem{ribeiro2016should}
Marco~Tulio Ribeiro, Sameer Singh, and Carlos Guestrin.
\newblock Why should i trust you? explaining the predictions of any classifier.
\newblock In {\em Proceedings of the 22nd ACM SIGKDD international conference
  on knowledge discovery and data mining}, pages 1135--1144, 2016.

\bibitem{Rice2020OverfittingIA}
Leslie Rice, Eric Wong, and J.~Zico Kolter.
\newblock Overfitting in adversarially robust deep learning.
\newblock In {\em ICML}, 2020.

\bibitem{salman}
Hadi Salman, Andrew Ilyas, Logan Engstrom, Ashish Kapoor, and Aleksander Madry.
\newblock Do adversarially robust imagenet models transfer better?
\newblock In H.~Larochelle, M.~Ranzato, R.~Hadsell, M.F. Balcan, and H.~Lin,
  editors, {\em Advances in Neural Information Processing Systems}, volume~33,
  pages 3533--3545. Curran Associates, Inc., 2020.

\bibitem{samangouei2018defensegan}
Pouya Samangouei, Maya Kabkab, and Rama Chellappa.
\newblock Defense-{GAN}: Protecting classifiers against adversarial attacks
  using generative models.
\newblock In {\em International Conference on Learning Representations}, 2018.

\bibitem{sehwag2022robust}
Vikash Sehwag, Saeed Mahloujifar, Tinashe Handina, Sihui Dai, Chong Xiang, Mung
  Chiang, and Prateek Mittal.
\newblock Robust learning meets generative models: Can proxy distributions
  improve adversarial robustness?
\newblock In {\em International Conference on Learning Representations}, 2022.

\bibitem{Shaham2015}
Uri Shaham, Yutaro Yamada, and Sahand Negahban.
\newblock Understanding adversarial training: Increasing local stability of
  supervised models through robust optimization.
\newblock {\em Neurocomput.}, 307(C):195–204, sep 2018.

\bibitem{song2018pixeldefend}
Yang Song, Taesup Kim, Sebastian Nowozin, Stefano Ermon, and Nate Kushman.
\newblock Pixeldefend: Leveraging generative models to understand and defend
  against adversarial examples.
\newblock In {\em International Conference on Learning Representations}, 2018.

\bibitem{szegedy2013intriguing}
Christian Szegedy, Wojciech Zaremba, Ilya Sutskever, Joan Bruna, Dumitru Erhan,
  Ian Goodfellow, and Rob Fergus.
\newblock Intriguing properties of neural networks.
\newblock {\em arXiv preprint arXiv:1312.6199}, 2013.

\bibitem{tramer2018ensemble}
Florian Tramèr, Alexey Kurakin, Nicolas Papernot, Ian Goodfellow, Dan Boneh,
  and Patrick McDaniel.
\newblock Ensemble adversarial training: Attacks and defenses.
\newblock In {\em International Conference on Learning Representations}, 2018.

\bibitem{Wong2020Fast}
Eric Wong, Leslie Rice, and J.~Zico Kolter.
\newblock Fast is better than free: Revisiting adversarial training.
\newblock In {\em International Conference on Learning Representations}, 2020.

\bibitem{wu2020adversarial}
Dongxian Wu, Shu-Tao Xia, and Yisen Wang.
\newblock Adversarial weight perturbation helps robust generalization.
\newblock In {\em Proceedings of the 34th International Conference on Neural
  Information Processing Systems}, NIPS'20, Red Hook, NY, USA, 2020. Curran
  Associates Inc.

\bibitem{xu2022safebench}
Chejian Xu, Wenhao Ding, Weijie Lyu, Zuxin Liu, Shuai Wang, Yihan He, Hanjiang
  Hu, Ding Zhao, and Bo~Li.
\newblock Safebench: A benchmarking platform for safety evaluation of
  autonomous vehicles.
\newblock In {\em Thirty-sixth Conference on Neural Information Processing
  Systems Datasets and Benchmarks Track}, 2022.

\bibitem{Yoon2021AdversarialPW}
Jongmin Yoon, Sung~Ju Hwang, and Juho Lee.
\newblock Adversarial purification with score-based generative models.
\newblock In {\em ICML}, 2021.

\bibitem{pmlr-v89-zhang19b}
Yuchen Zhang and Percy Liang.
\newblock Defending against whitebox adversarial attacks via randomized
  discretization.
\newblock In Kamalika Chaudhuri and Masashi Sugiyama, editors, {\em Proceedings
  of the Twenty-Second International Conference on Artificial Intelligence and
  Statistics}, volume~89 of {\em Proceedings of Machine Learning Research},
  pages 684--693. PMLR, 16--18 Apr 2019.

\end{thebibliography}








\end{document}